\documentclass[conference]{IEEEtran}
\IEEEoverridecommandlockouts
\usepackage{cite}
\usepackage{booktabs} 
\usepackage{multirow}
\usepackage{pgfplots}
\usepackage{subfig}
\usepackage{array}
\usepackage{siunitx}
\usepackage[all]{nowidow}

\pgfplotsset{compat=1.9}
\usetikzlibrary{shadows,patterns,shapes,arrows,decorations.pathmorphing,backgrounds,positioning,fit,plotmarks,calc,spy}
\usepgfplotslibrary{patchplots}
\usepackage{todonotes}

\usetikzlibrary{pgfplots.statistics}
\usetikzlibrary{snakes}
\usepackage{amsmath,amssymb,latexsym}
\usepackage{textcomp}
\usepackage[all]{nowidow}
\usepackage{gensymb}
\usepackage{amsthm}
\usepackage{color}
\usepackage{algorithm}
\usepackage{algpseudocode}

\pgfplotsset{ /pgfplots/ybar legend/.style={
		/pgfplots/legend image code/.code={
			\draw[##1,/tikz/.cd,bar width=3pt,yshift=-0.2em,bar shift=0pt]
			plot coordinates {(0cm,0.8em)};},
	},
}

\begin{document}

\title{Neural Network Inference on Mobile SoCs}

\author{ 
	Siqi~Wang,
	Anuj~Pathania,
	Tulika~Mitra
		\thanks{ S. Wang, A. Pathania and T. Mitra are with the Department of Computer Science, School of Computing, National University of Singapore. E-mail:~((wangsq, pathania, tulika)@comp.nus.edu.sg). Address: COM1, 13 Computing Drive, S117417.
		\protect (Corresponding author: Tulika Mitra)\protect\\
	}
}

\maketitle

\begin{abstract}
	
The ever-increasing demand from mobile Machine Learning~(ML) applications calls for evermore powerful on-chip computing resources.
Mobile devices are empowered with heterogeneous multi-processor Systems-on-Chips~(SoCs) to process ML workloads such as Convolutional Neural Network~(CNN) inference.  Mobile SoCs house several different types of ML capable components on-die, such as CPU, GPU, and accelerators. 
These different components are capable of independently performing inference but with very different power-performance characteristics. 
In this article, we provide a quantitative evaluation of the inference capabilities of the different components on mobile SoCs. 
We also present insights behind their respective power-performance behavior. 
Finally, we explore the performance limit of the mobile SoCs by synergistically engaging all the components concurrently. 
We observe that a mobile SoC provides up to 2x improvement with parallel inference when all its components are engaged, as opposed to engaging only one component.

\end{abstract}


\section {Introduction} \label{Introduction}

The tremendous popularity of Neural-Network (NN) based machine learning applications in recent years has been fuelled partly by the increased capability of the compute engines, in particular, the GPUs. Traditionally, both the network training and inference were performed on the cloud with mobile devices only acting as user interfaces. However, enriched user experience and privacy concerns now demand inference to be performed on the mobile devices themselves with high accuracy and throughput. 

In this article, we look at NN-enabled vision applications on mobile devices. These applications extract high-level semantic information from real-time video streams and predominately use Convolutional Neural Networks~(CNNs). They are important in many domains, such as Advanced Driver-Assistance Systems~(ADAS), Virtual Reality~(VR), and Augmented Reality~(AR). Enabling these applications in the power-constrained mobile devices is challenging due to the enormous computational and memory requirements.

Heterogeneous multi-processor SoC enables the current state-of-the-art mobile devices. 
However, the presence of multiple vendors fragments the mobile SoCs. 
Accelerators (including GPU, FPGA, and dedicated neural accelerators) demonstrate great performance for inference. 
However, these high-performance components are present in only a small fraction of the mobile devices. 
Moreover, due to market fragmentation, it is impossible to develop a mobile application with accelerators that can run across multiple devices. Instead, the CPUs remain the common denominator among mobile SoCs and is the favored choice for inference~\cite{wu2019machine}.

We embark on an exploration to quantitatively characterize and understand the inferencing capabilities of the mobile SoCs given the diverse landscape. We portray the power-performance gap between the ubiquitous CPUs and the high-performance accelerators in high-end devices and uncover the reasons behind the gap through the roofline models. Finally, we propose simultaneous engagement of all the SoC components to greatly expand the promise of functional deployment of vision applications on mobile devices.

\begin{figure}[!t]
	\centering
	\scriptsize
	\sffamily
	\begin{tikzpicture}[x=8.5, y=9]
	
	\draw[fill=white, rounded corners] (0,3) rectangle +(9,10.5);
	\draw[fill=white, rounded corners] (.25,9.5) rectangle +(4,2) node[pos=.5,] {Core};
	\draw[fill=white, rounded corners] (4.75,9.5) rectangle +(4,2) node[pos=.5,] {Core};	
	\draw[fill=white, rounded corners] (.25,6.5) rectangle +(4,2) node[pos=.5,] {Core};
	\draw[fill=white, rounded corners] (4.75,6.5) rectangle +(4,2) node[pos=.5,] {Core};
	\draw[fill=white, rounded corners] (0.5,3.5) rectangle +(8,2) node[pos=.5,] {L2 Cache};
	
	\draw[fill=white, rounded corners] (10,3) rectangle +(8,9.5);
	\draw[fill=white, rounded corners] (10.25,8.5) rectangle +(3.5,2) node[pos=.5,] {Core};
	\draw[fill=white, rounded corners] (14.25,8.5) rectangle +(3.5,2) node[pos=.5,] {Core};
	\draw[fill=white, rounded corners] (10.25,6) rectangle +(3.5,2) node[pos=.5,] {Core};
	\draw[fill=white, rounded corners] (14.25,6) rectangle +(3.5,2) node[pos=.5,] {Core};
	\draw[fill=white, rounded corners] (11,3.5) rectangle +(6,2) node[pos=.5,] {L2 Cache};
	
	\draw[fill=white, rounded corners] (19,7) rectangle +(9,5)  node[pos=.5,] {GPU};

	\draw[fill=white, rounded corners] (22,3.5) rectangle +(3,2) node[pos=.5,] {NPU};
	
	\draw[fill=white, rounded corners] (0,1.5) rectangle +(28,1) node[pos=.5,] {CCI Bus};
	
	\draw[fill=white] (0,-2) rectangle +(28,3) node[pos=.5,] {DRAM};
	
	\draw[fill=white] (5.5,12) rectangle +(3,1) node[pos=.5,] {DVFS};
	\draw[fill=white] (24.5,10.5) rectangle +(3,1) node[pos=.5,] {DVFS};
	\draw[fill=white] (14.5,11) rectangle +(3,1) node[pos=.5,] {DVFS};
	
	\draw[white] (0.25,12) rectangle +(4,1) node[pos=.5,black] {Big CPU};
	\draw[white] (10.25,11) rectangle +(4,1) node[pos=.5,black] {\tiny Small CPU};
	
	\draw[dashed, rounded corners] (19,3) rectangle +(9,3) node[pos=.5,black] {};
	
	\end{tikzpicture}
	\caption{An abstract block diagram of a mobile SoC with an asymmetric multi-core CPU, GPU, and NPU.}
	\label{fig:block}
\end{figure}
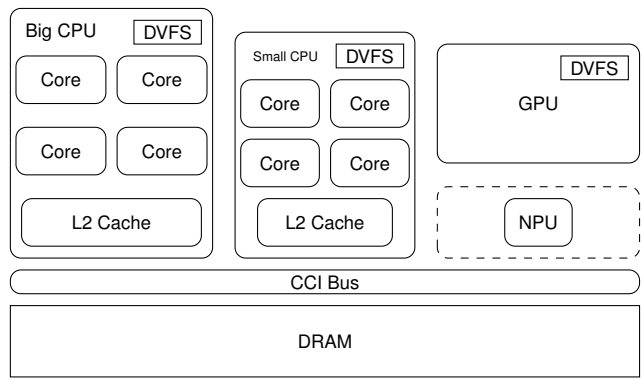

\section{Inference on Mobile SoCs}

\subsection{Heterogeneous Multi-processor SoCs}
There are over two thousand unique mobile SoCs in the mobile devices market. The diversity comes from the choice of different CPUs, GPUs, caches, memory controllers, and other application-specific accelerators. This fragmentation of the SoC market makes standard optimizations impossible. However, the similarity among these SoCs lies in the choice of one or more CPU core clusters.

\subsubsection{ARM big.LITTLE}
Multi-cores enable the state-of-the-art Mobile SoCs. 
99.9\% of the {\em Android} devices in the market in 2019 have multiple cores~\cite{wu2019machine}. Among these, about half of the SoCs implement performance heterogeneity with at least two CPU clusters: a high-performance and an energy-efficient core cluster. {\em ARM big.LITTLE} architecture, one of the most popular architectures implementing this heterogeneity, is present in \textit{Hi-Silicon Kirin}, \textit{Samsung Exynos}, and \textit{Qualcomm Snapdragon} series SoCs. The heterogeneous cores differ in power-performance-area characteristics but share the same Instruction Set Architecture~(ISA). Figure~\ref{fig:block} shows an abstract block diagram of this architecture. The general availability of CPUs make them a favorable choice for mobile inference and make device-agnostic optimizations feasible.

\subsubsection{Accelerators}
Existing architectures, including GPU and FPGA, have proven to be advantageous for ML workloads and are thus commonly used for deployment on certain devices. Both academic and commercial dedicated accelerators (\textit{Google Edge TPU}, \textit{Intel Nervana NNP}, \textit{Huawei NPU}, \textit{Apple Neural Engine}) offer exceptional runtime and energy-efficiency. There are no standard neural accelerators for mobile SoCs, making horizontal application integration difficult. 
Limited availability even constraints the use of GPUs. 

\subsection{Mobile ML Framework and Optimizations}

\textit{Tensorflow}, \textit{PyTorch}, and \textit{MXNet} are some of the common ML development frameworks for all scenarios. \textit{Tensorflow Lite} like frameworks facilitates the compression of huge models to fit into resource-constrained mobile devices. Efficient libraries and APIs bridge the gap between the frameworks and the underlying hardware, examples of which are \textit{Nvidia cuDNN} for GPUs, \textit{ARM NN} powered by \textit{Compute Library (ARM-CL)} for ARM CPUs and GPUs, \textit{Facebook} \textit{NNPACK}, and \textit{QNNPACK} for mobile CPUs. These libraries usually optimize with detailed architectural information. {\em ARM-CL} supports acceleration through {\em ARM} NEON vectorization and provides NEON assembly implementation for the most computationally intensive convolution kernels. Algorithmic optimizations (Winograd transform, FFT, sparsity exploration) lower the computational complexity of convolution computations. Furthermore, quantization and network pruning are common techniques that bring down the processing requirement with the sacrifice of accuracy~\cite{wess2018weighted}. 

Even though most mobile inference workloads run on CPUs, optimizations of ML workloads with accelerators hordes most of the attention. There is a lot of room for optimizations on mobile CPUs to enable ML applications across different mobile platforms.

\section{Characterizing Inferencing on Mobile SoC}
We perform experiments across different technology nodes using two commonly used mobile SoCs:  28\,nm {\em Exynos 5422} within {\em Odroid XU3} development platform and 10\,nm {\em Kirin 970} within {\em Hikey 970} development platform. 
Released in 2014 and 2017 respectively, these two SoCs show us the progress of mobile SoCs development over the years. Furthermore, these two SoCs roughly approximate the mid- and high-end mobile SoCs today.

In the experiments, both SoCs are using {\em ARM-CL 18.05v}. {\em Kirin 970} NPU is supported by {\em HiAI DDK (v100)} for network deployment. For \textit{Exynos5422}, in-built power sensors, running at 200\,Hz, measure the power of each component. For {\em Kirin 970}, because of the absence of any integrated on-chip power sensors, we approximate the power consumption by measuring the socket power with the help of a power measurement unit~\cite{pmu} running at 100\,Hz.

\subsection{Experimental Set-up}
\subsubsection{CPU}
Both SoCs include {\em ARM big.LITTLE} based asymmetric multi-core CPU. {\em Kirin 970} CPU adopts {\em ARMv8-A} architecture. It consists of a high-performance high-power out-of-order four-core {\em Cortex-A73} cluster (2.36\,GHz) and a low-performance low-power four-core in-order {\em Cortex-A53}~(1.8\,GHz).
{\em Exynos 5422} has a similar design but uses an older {\em ARMv7-A} architecture with {\em Cortex-A15}~(2\,GHz) and {\em Cortex-A7}~(1.4\,GHz) cores. 
All CPU cores support NEON advanced Single Instruction Multiple Data~(SIMD) operations, which allows for four 32-bit floating-point operations per cycle. 
		
\subsubsection{GPU}
{\em Kirin 970} adopts {\em ARM Mali G72 MP12 GPU} (850\,MHz), implementing the second generation {\em Bifrost} architecture. It has twelve shader cores with three execution engines each. 
Each engine is capable of eight FP32 operations per cycle, giving a total peak compute capability of 244.8\,GFLOPS/s for {\em G72}. 
{\em Exynos 5422} includes an {\em ARM Mali T628 MP6} GPU (600\,MHz). It adopts an older {\em Midgard} architecture with six shader cores implementing {\em Tripipe} design with two arithmetic pipelines.
Each pipeline is capable of eight FP32 operations per cycle, providing a total peak compute capability of 57.6\,GFLOPS/s for {\em T628}.
				
\subsubsection{NPU}
{\em Kirin 970} includes a {\em Huawei} NPU purpose-built for ML. 
It has a peak performance of 1.92\,TFLOPS/s with FP16.
The accompanying {\em HiAi DDK} API  enables the deployment of networks on NPU but only works with {\em Android}. {\em Exynos 5422} does not have any ML accelerator.

\subsubsection{Network Structure}
We experiment with several popular networks introduced in recent years -- {\em AlexNet}~\cite{alexnet}, {\em GoogleNet}~\cite{googlenet}, {\em MobileNet}~\cite{mobilenet}, {\em ResNet50}~\cite{resnet}, and {\em SqueezeNet}~\cite{squeezenet}.

\subsection{Individual Heterogeneous Components}

\begin{table}[!t]
	\caption{Throughput of different networks on different mobile SoCs components running at their peak frequencies.}
		\label{tab:HMPSOC_Comparision}
		\centering
			\resizebox{\columnwidth}{!}{
	\begin{tabular}{l|r|r|r|r|r|r|r}
		\hline
		\multirow{2}{*}{Network} & \multicolumn{3}{c|}{\begin{tabular}[c]{@{}c@{}}\textit{Exynos 5422}\\ Throughput (Imgs/s)\end{tabular}} & \multicolumn{4}{c}{\begin{tabular}[c]{@{}c@{}}\textit{Kirin 970}\\ Throughput (Imgs/s)\end{tabular}} \\ \cline{2-8} 
		& \multicolumn{1}{c|}{\textit{A7}} & \multicolumn{1}{c|}{\textit{A15}} & \multicolumn{1}{c|}{\textit{T628}} & \multicolumn{1}{c|}{\textit{A53}} & \multicolumn{1}{c|}{\textit{A73}} & \multicolumn{1}{c|}{\textit{G72}} & \multicolumn{1}{c}{NPU} \\ \hline
		\textit{AlexNet} & 1.1 & 3.1 & 7.8 & 2.2 & 7.6 & 32.5 & 32.5 \\ \hline
		\textit{GoogLeNet} & 0.9 & 3.4 & 5.2 & 3.0 & 7.1 & 19.9 & 34.4 \\ \hline
		\textit{MobileNet} & 1.5 & 5.7 & 8.5 & 6.5 & 17.7 & 29.1 & \multicolumn{1}{c}{\scriptsize \begin{tabular}[c]{@{}c@{}} \textit{Not} \textit{Supported}\end{tabular}} \\ \hline
		\textit{ResNet50} & 0.2 & 1.3 & 2.1 & 1.5 & 2.8 & 8.4 & 21.9 \\ \hline
		\textit{SqueezeNet} & 1.5 & 5.0 & 8.0 & 6.8 & 15.7 & 43.0 & 49.3 \\ \hline
	\end{tabular}
}
\end{table}

We first study each component in isolation by running inferencing of multiple images in a stream on a single component.
Both {\em Big} and {\em Small} clusters are self-sufficient for inferencing.
GPU and NPU require the support of a {\em Small} cluster for inferencing.

\subsubsection{Throughput}
Table~\ref{tab:HMPSOC_Comparision} shows the throughput of each component on both our SoCs. All components in {\em Kirin 970} outperform their respective counterparts in older {\em Exynos 5422}. {\em Big A73} cluster, {\em Small A53} cluster, and {\em G72} GPU outperform {\em Big A15} cluster, {\em Small A7 cluster}, and {\em T628} GPU on average by a factor of 4.4x, 2.6x, and 4.2x, respectively. The performance gap between the {\em Big} and {\em Small}  cluster has reduced from 4x to 2.5x with a decrease in {\em Big} to {\em Small} power consumption ratio from 10x to 4x. Furthermore, the performance gap between GPU and CPU clusters is only about 2x to 3x for both SoCs.

For NPU, we were unable to deploy {\em MobileNet} due to incompatible operators. On average, NPU is only 1.6x better than the high-end {\em G72} GPU.
On the other hand, the portability of applications across different platforms remains a challenge for dedicated accelerators.
The proprietary development kit makes the general optimization a difficult endeavor.

\subsubsection{Energy Efficiency}

\begin{figure} [!t]
	\centering
\includegraphics[width=0.98\columnwidth]{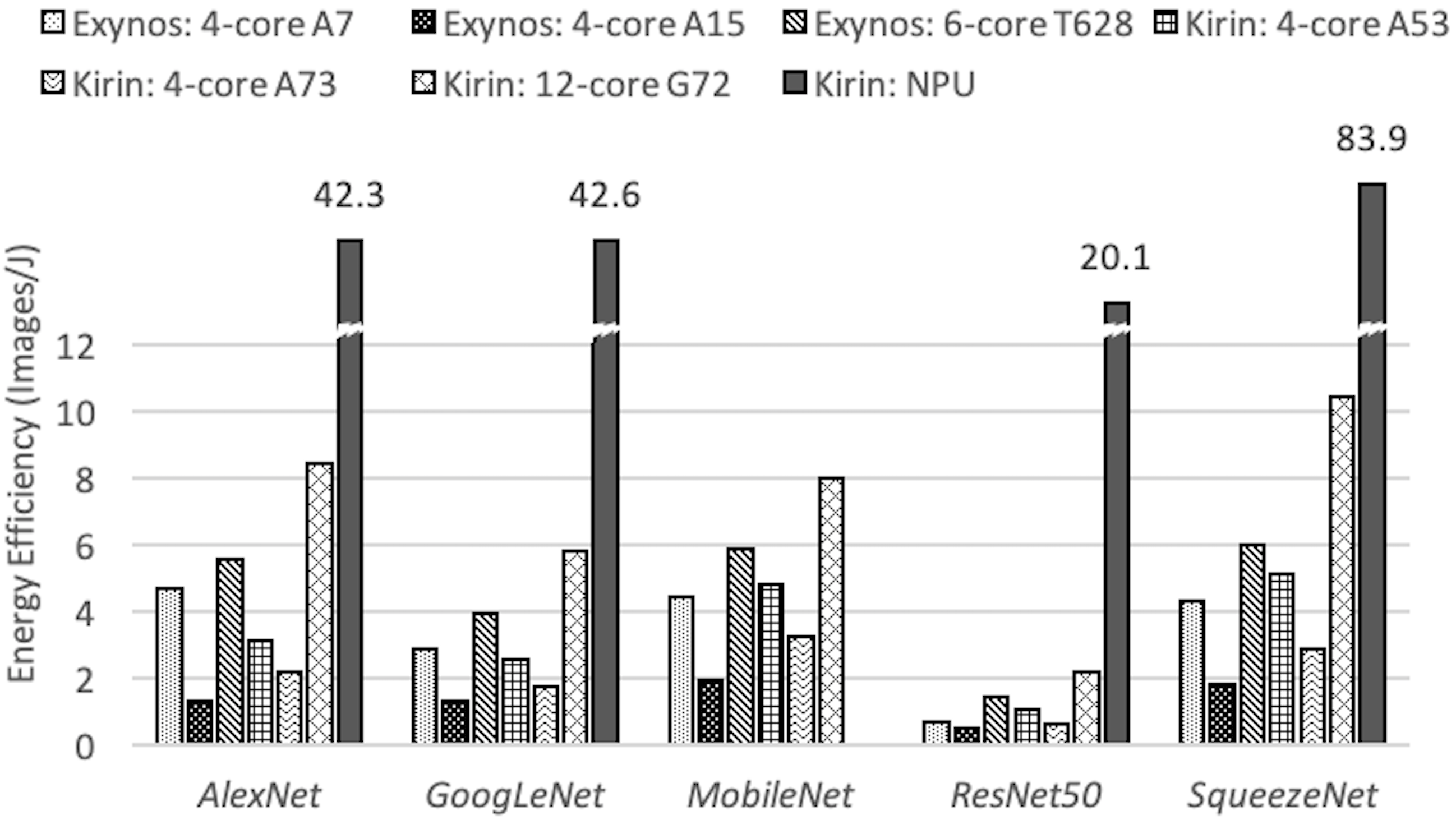}
\caption{Energy efficiency of different components while running at their peak frequencies.}
\label{fig:power_efficiency}

\end{figure}

We measure the average active power consumption of inferencing on different components and calculate the energy efficiency, as shown in 
Figure~\ref{fig:power_efficiency}.
For \textit{Exynos 5422}, power sensors for individual components measure the power consumption of each component separately. For \textit{Kirin 970}, we calculate active power values by subtracting the idle power (measured when no workload is running) from socket power measurement taken during inferencing. Therefore, the power measurements for \textit{Kirin} are slightly higher, as memory power cannot be separated.

NPU is the most energy-efficient among all components, which we expect, given its custom design for inference.
GPUs are the second-most energy-efficient component.
{\em Small} clusters also show good energy-efficiency. 
However, Table~\ref{tab:HMPSOC_Comparision} shows their performance in terms of absolute throughput is too low to be ever useful alone. 

Comparing across two platforms, the energy efficiency of each component has improved for the newer SoC. However, the improvement is minimal and even negative for the \textit{Small} CPU cluster. Compared to its predecessor A7, A53 is more complex and area hungry with 64-bit, complex branch prediction, and larger TLB.
It achieves greater performance but at the cost of even greater power consumption.

\subsubsection{Impact of Technology Scaling Versus Architectural Innovations}
{\em Exynos 5422} and {\em Kirin 970} use the 28\,nm and 10\,nm technology nodes, respectively. In moving from 28\,nm {\em Exynos 5422} to 10\,nm {\em Kirin 970}, the maximum frequency of the {\em Big} cluster has only changed from 2\,GHz ({\em A15}) to 2.36\,GHz ({\em A73}), while the {\em Small} cluster changes from 1.4\,GHz ({\em A7}) to 1.8\,GHz ({\em A53}). So the frequency scaling is 1.18x for the big cluster and 1.29x for the {\em Small} cluster for these two platforms. On the other hand, we get 4.4x and 2.6x throughput improvement across technology generations (Table~\ref{tab:HMPSOC_Comparision}) for {\em Big} cluster and {\em Small} cluster, respectively. This improvement in performance is achieved through smart designs such as micro-architectural improvements (improved branch predictor, cache data prefetchers, etc.), larger caches, and 64-bit support leading to improved NEON processing, among others.
	
However, in the case of the small cluster, with an increased area, the micro-architectural changes give an increase in power that cannot be offset by technology scaling. Indeed, the small {\em A53} cluster consumes roughly twice the power of the small {\em A7} cluster. Thus, the energy-efficiency improvement is limited for the small cluster for some networks as we move from {\em A7} to {\em A53}. In contrast, between the two big clusters, {\em A73} is more power-efficient compared to {\em A15}; the energy-efficiency improves from {\em A15} to {\em A73} cluster. As mentioned earlier, the power measurements for {\em A7} and {\em A15} are quite accurate, while the measured power for {\em A53} and {\em A73} are higher as it includes the memory power that could not be separated.

\subsubsection{Insights}

We observe that NPU provides unmatched energy-efficiency for inferences. It is the optimal choice to perform network inferences on the platforms with such dedicated accelerators. However, a developer needs to put in substantial effort to port their application with proprietary API to execute on NPU, and the effort would not bear any fruits on mobile devices lacking this very-specific NPU. NPU, as a black-box, also causes inflexibility in development and optimizations.
Furthermore, NPU is compatible with only a limited set of network designs. These extra requirements could make it quickly obsolete for future networks.

On the other hand, high-end GPUs can provide performance comparable to NPU at satisfactory energy-efficiency.
GPUs are capable of running General-Purpose (GPGPU) applications written in {\em OpenCL}, which is easily portable to a large variety of GPUs and even CPUs supporting \textit{OpenCL}. This generality makes it a good candidate to use when high performance is a major consideration.

CPUs provide both the worst energy-efficiency as well as the worst throughput among all components. 
Still, they are critical for inferencing because they are commonly present across all mobile devices.
Low-end mobile SoCs would lack accelerators like NPU. They may contain a low-end GPU, but maybe missing {\em OpenCL} support and thereby lack any inferencing capability. Network inference on CPU is inevitable and demands optimization considerations.

Our analysis shows that any component alone on both platforms can barely support the increasing performance requirement for network inferencing. Section~\ref{Sec:Co-Execution} presents the co-execution methodology that can mitigate the performance issue to some extent. Still, we must continue to look into the networks themselves in search of further optimization opportunities.

\section{Roofline Analysis}
To understand the execution behaviors of the networks on each SoC components, we perform a roofline analysis. Roofline analysis~\cite{roofline} is a widely applied methodology that can classify an application as memory- or compute-bound on given hardware. It gives insights to developers for improving their application design to cater to the computation and memory capability of the underlying processing devices. The horizontal ``Ceiling'' and the ``Roof'' constructs a ``Roofline'' that bounds the maximum performance of an application~(measured in GOPS/s) under a hardware-determined compute- or memory-bound, respectively. Operational Intensity~(OI) of application~(measured in FLOPS/byte) determines whether its peak performance is bounded by the memory bandwidth~(measured in GB/s) or compute capability~(measured in GOP/s) of the hardware.
Both {\em Exynos 5422} and {\em Kirin 970} show similar behavior for the CPU core clusters and GPU.
Therefore, we only present here the analysis for {\em Exynos 5422}. 

\subsection{Construction of a Roofline Model}

\begin{figure*}[t!]
	\centering	
	\includegraphics[width=0.95\textwidth]{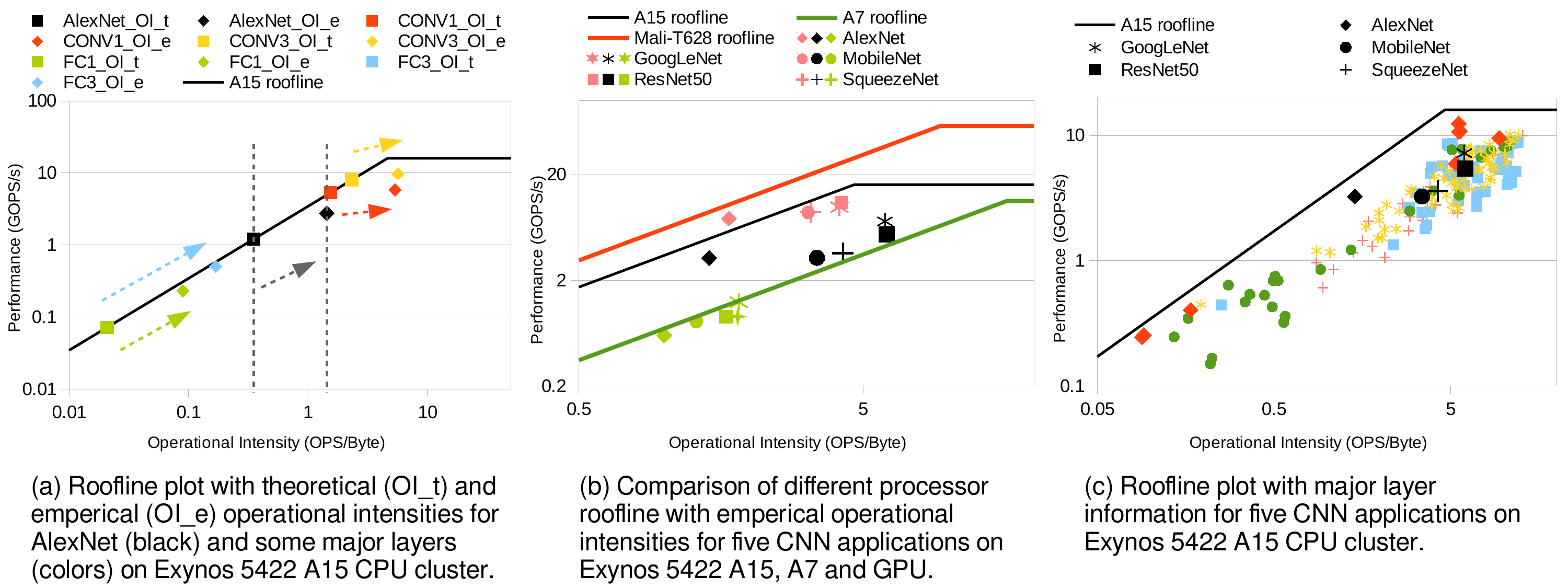}
	\caption{Roofline plot for inference workloads and major layer information on multiple processors in {\em Exynos 5422}.}
	\label{fig:rl}
\end{figure*}

Hardware specifications provide the peak pure compute performance. Micro-benchmarking~\cite{tinymembench} provides the peak (sustainable) memory bandwidth. Specifications claim peak memory bandwidth of the memory bus to be 14.9\,GB/s. However, we observe the actual component-wise peak bandwidth to be 3.44\,GB/s, 0.49\,GB/s, and 6.15\,GB/s for {\em A15} cluster, {\em A7} cluster, and {\em T628} GPU, respectively.

Many variations of the roofline model are constructed to adapt to different use-cases. In this analysis, we defined two operational intensities, that are, theoretical OI ($OI_t$) and empirical OI ($OI_e$), defined in Eqn~\eqref{oit} and \eqref{oie}.
\begin{equation}\label{oit}
OI_t = GOPS / Mem\_Access
\end{equation}
\begin{equation}\label{oie}
OI_e = GOPS / DRAM\_Access
\end{equation}
We calculate $OI_t$ by analyzing the code. The memory accesses include all the data required in the computation. During actual executions, multiple levels of caches within components improve the memory access performance. The caches make it difficult for $OI_t$ to correlate with the actual performance on the components. Therefore, we introduce empirical operational intensity $OI_e$.
We calculate $OI_e$ using the actual DRAM accesses on the bus, which models the presence of multi-level memory hierarchy. It is more informative and has a better correlation with the actual performance on the component than $OI_t$. We use application-specific performance counters obtained from {\em ARM Streamline DS5} at run-time for calculation of $OI_e$ (CPU: \textit{L2\_data\_refill}, GPU: \textit{Mali L2 cache external read/write bytes}). Fig.~\ref{fig:rl}(a) show the roofline points of major layers in {\em AlexNet} on {\em A15} cluster for both $OI_t$ and $OI_e$.

\subsection{Theoretical and Empirical OI}
Figure~\ref{fig:rl}(a) plots the $OI_t$ (squares) and $OI_e$ (diamonds) values of several {\em AlexNet} major layers, marked with different colors. 
Black marks the whole network $OI_t$ and $OI_e$  of AlexNet. The intersection points of the $OI_t$ values with the ``Roofline'' represent the theoretical maximum performance for the code-based theoretical operational intensities, which fall in the memory-bound region on the ``Roof''. The corresponding points for $OI_e$ are actual achieved performance in GOPS/s, which are always below the ``Roofline''.

The presence of cache reduces the memory accesses going to the DRAM during execution, and thus increases the operational intensity. Therefore, for all layers, $OI_e$ points are on the right of $OI_t$ points, indicating higher performance. For layers with low $OI_t$ (fully connected, FC), the  points move along the ``Roofline'', achieving the theoretical maximum performance. For layers with higher $OI_t$ (convolutional, CONV), the points cross the boundary of memory-bound and become compute-bound. The performance gain is not as significant, and we explain this with the underutilization due to insufficient or imperfect parallelization. Overall, $OI_e$ is a better indicator of real-world performance. Therefore, we only plot values of $OI_e$ going forward.

\subsection{Across Different Components}
Figure~\ref{fig:rl}(b) shows the performance of different networks on different components on {\em Exynos 5422}. The color of the points corresponds to the respective component. We can observe that memory severely bottlenecks the performance of both {\em A7} cluster and {\em T628} GPU. Performance of {\em A15} cluster falls in both compute- and memory-bound regions depending upon the network.

The $OI_e$ values are different because of the different memory hierarchies for different components. The {\em Big} core cluster with a larger cache size (L2: 2MB) derives higher benefits from memory hierarchy than GPU (L2: 128KB). However, \textit{AlexNet} that is notorious for huge parameter sizes caches will get flushed regardless of the cache sizes resulting in a smaller benefit from the memory hierarchy. On the other hand, small filter sizes lead to sub-optimal parallelization (under-utilization). This observation holds more starkly for newer networks with smaller filter size than older networks. The observation explains the significant deviation in the empirical performance of networks on the components from the ``Roofline''.

\subsection{Major Layers in Inference}
 
We do a deeper layer-level analysis to explain the behavior of the networks. Both convolutional and fully-connected layers dominate the total execution time of networks, and thus both are considered as major layers worthy of examination. We limit our analysis to {\em Big} cluster because networks there show both memory- and compute-bound behavior. Figure~\ref{fig:rl}(c) shows that different layers in {\em AlexNet}~(and also other networks to a lesser extent) exhibits different empirical OIs. Convolutional layers at the start of {\em AlexNet}  perform compute-intensive convolution on large inputs and thereby have relatively higher OIs. On the other hand, fully-connected layers perform memory-intensive operations on large size parameters and thereby have relatively lower OIs. Convolutional and fully-connected layers of {\em AlexNet} fall in the compute- and memory-bound region of the roofline model, respectively. Overall, {\em AlexNet} falls somewhere in the middle of both.

In general, we observe that layers of a network are scattered in both compute- or memory-bound region. This difference comes from the choice of the size of the input tensors and filters. The vast differences in $OI_e$ for different layers within a network motivates layer-level optimizations such as per-layer Dynamic Voltage and Frequency Scaling~(DVFS) for power management. Furthermore, the variation within a network motivates fine-grain layer level co-executions, which improve the overall chip utilization~\cite{pipe-it}.

\subsection{Effect of Quantization}
Quantization is a commonly applied technique that reduces the memory and computation requirement of a network while reducing accuracy. 
However, the quality of its implementation primarily determines the benefits it provides.
In the implementation of quantized MobileNet in {\em ARM-CL} (18.05v), QASYMM8 model with 8-bit weights is used. This implementation fails to improve the overall performance of the network. Deeper analysis reveals that the latencies of convolutional layers are indeed reduced, but the overheads from extensive de-quantization and re-quantization overshadow any benefit.

Quantization reduces the total operations and memory access required near-proportionally. 
Reduction in  memory accesses results in a slightly higher empirical operational intensity $OI_e$. 
Therefore, the roofline analysis of a quantized network nearly overlaps with that of its non-quantized counterpart, and quantization does not improve the memory behavior of the layers.
Lower operation requirements under quantization predominately contribute to the reduction in execution time of the convolutional layers.

\subsection{Glimpse of NPU}
NPU, due to its novelty and dedicated machine learning processing design, garners a lot of attention. 
However, most of the details are kept confidential.
We are unaware of its architectural  and integration details. 
Therefore, we can only attempt to reverse engineer its behavior to gain some insights.

We implement a kernel module that enables counting of traffic on the CCI bus. We attribute the traffic on the CCI bus that goes to DRAM during the engagement of NPU to the main memory activity of NPU. The maximum observed memory bandwidth of executing several networks and the peak performance of 1.92 TOPS from the specification construct the ``Roof'' and ``Ceiling'' of the NPU roofline. 
We observe that the performance of NPU is significantly bounded by the memory for the networks tested. This observation shows a significant scope for optimization to achieve the full processing potential of NPU.

\section{Improving the performance}

\subsection{Co-Execution of Multiple Components} \label {Sec:Co-Execution}
\begin{table}[!t]
	\caption{Throughput improvement on {\em Exynos 5422} and {\em Hikey 970}  by co-execution over the best throughput with a single component~({\em T628} and {\em G72} GPU).}
	\label{tab:odroid_res}
	\resizebox{\columnwidth}{!}{
\begin{tabular}{l|r|r|r|r|r|r}
	\hline
	\multirow{2}{*}{Network} & \multicolumn{3}{c|}{\begin{tabular}[c]{@{}c@{}}\textit{Exynos 5422} \\ Throughput (Imgs/s)\end{tabular}} & \multicolumn{3}{c}{\begin{tabular}[c]{@{}c@{}}\textit{Kirin 970} \\ Throughput (Imgs/s)\end{tabular}} \\ \cline{2-7} 
	& \multicolumn{1}{c|}{\textit{T628}} & \multicolumn{1}{c|}{\begin{tabular}[c]{@{}c@{}}Co-\\ execution\end{tabular}} & \multicolumn{1}{c|}{Gain} & \multicolumn{1}{c|}{\textit{G72}} & \multicolumn{1}{c|}{\begin{tabular}[c]{@{}c@{}}Co-\\ execution\end{tabular}} & \multicolumn{1}{c}{Gain} \\ \hline
	\textit{AlexNet} & 7.8 & 10.3 & 32.4\% & 32.5 & 33.4 & 2.8\% \\ \hline
	\textit{GoogLeNet} & 5.2 & 8.7 & 66.3\% & 19.9 & 28.4 & 42.8\% \\ \hline
	\textit{MobileNet} & 8.5 & 14.9 & 76.7\% & 29.1& 51.5& 77.1\% \\ \hline
	\textit{ResNet50} & 2.1 & 2.9 & 38.6\% & 8.4 & 12.3& 46.3\% \\ \hline
	\textit{SqueezeNet} & 8.0 & 13.8 & 73.9\% & 43.0 & 54.5 & 26.7\% \\ \hline
\end{tabular}
	}
\end{table}

Stream processing, depending on the application, requires  10 to 40 images/second throughput. Some applications even require multiple inferences to run at the same time.
Table~\ref{tab:HMPSOC_Comparision} shows that the high-end \textit{Kirin 970} SoC can barely sustain such requirement while the mid-end \textit{Exynos 5422} cannot.
We previously observed that peak bandwidth consumed by any individual component is far below the total bandwidth supported by the bus. This observation supports the claim that inferencing through multiple components together will not make individual components more memory-constrained compared to their isolated inferencing.
Therefore, we use {\em ARM-CL} to create an infrastructure, wherein multiple components process images from a single unified stream in parallel using a work-stealing mechanism. The infrastructure uses a buffer to reorder the out-of-sync output from different components.
Co-execution obtains significantly higher throughput than the highest throughput component in isolated execution. 

Table~\ref{tab:odroid_res} shows the peak co-execution throughput on both mobile SoCs with the \textit{ARM big.LITTLE} CPU core cluster and GPU. We include the best individual component executions, which are GPU for both platforms, for comparison. On average, the co-execution gives 50\% throughput improvement over GPU only execution. 
Furthermore, Table~\ref{tab:odroid_res} shows {\em Exynos 5422}'s obsolescence. Even with the co-execution, {\em Exynos 5422} shows very low absolute throughput.  

\subsection{Co-execution with NPU}
\begin{table}[!t]
	\caption{Throughput improvement on {\em Kirin 970} by co-execution over the best throughput with a single component~(NPU).}
	\label{tab:hikey_res}
	\resizebox{\columnwidth}{!}{
		\begin{tabular}{l|r|r|r|r|r|r|l}
			\hline
			\multicolumn{1}{c|}{\multirow{2}{*}{Network}} & \multicolumn{2}{c|}{Throughput (Images/s)} & \multicolumn{1}{c|}{\multirow{2}{*}{\begin{tabular}[c]{@{}c@{}}Gain\\ (\%)\end{tabular}}} & \multicolumn{4}{c}{\begin{tabular}[c]{@{}c@{}}Image Frames \\ Composition (\%)\end{tabular}} \\ \cline{2-3} \cline{5-8} 
			\multicolumn{1}{c|}{} & \multicolumn{1}{c|}{\begin{tabular}[c]{@{}c@{}}NPU\end{tabular}} & \multicolumn{1}{c|}{\begin{tabular}[c]{@{}c@{}}Co-\\ execution\end{tabular}} & \multicolumn{1}{c|}{} & \multicolumn{1}{c|}{\em A73} & \multicolumn{1}{c|}{\em A53} & \multicolumn{1}{c|}{\em G72} & NPU \\ \hline
			{\em AlexNet} & 32.5 & 63.7 & 96.0 & 1.90 & 0.95 & 47.47 & 49.68 \\ \hline
			{\em GoogleNet} & 34.4 & 59.3 & 72.4 & 3.06 & 1.70 & 33.33 & 61.90 \\ \hline
			{\em ResNet50} & 21.9 & 30.9 & 40.9 & 2.63 & 1.32 & 26.97 & 69.08 \\ \hline
			{\em SqueezeNet} & 49.3 & 95.1 & 92.9 & 3.18 & 1.69 & 43.43 & 51.69 \\ \hline
		\end{tabular}
	}
\end{table} 

The performance of NPU is unbeatable. Table~\ref{tab:hikey_res} shows that {\em Kirin 970}, with co-execution of all on-chip components, gives exceptionally high throughput. In practice, we can execute NPU and GPU  in parallel towards one application that demands very high performance or to perform multiple inferences simultaneously with multiple applications.

\subsection{Co-Execution Energy Efficiency}
\begin{figure}[!t]
		\vspace{-0.1in}
	\centering	\includegraphics[width=\columnwidth]{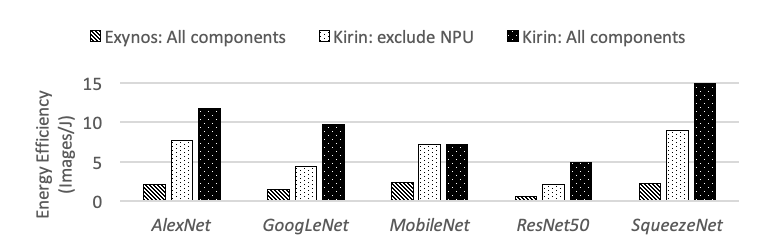}
	\caption{Energy efficiency of co-execution on \textit{Exynos 5422} with all components, on \textit{Kirin 970} with CPU and GPU~(excluding NPU) and all components~(including NPU).}
	\label{fig:ee-coexe}
\end{figure}

Synergistic co-execution engages multiple components simultaneously to improve performance at the cost of higher power consumption. 
Therefore, the energy efficiency of the co-execution is the average energy efficiency of engaged components.
Figure~\ref{fig:ee-coexe} shows the energy efficiency of the execution that engages all the components on \textit{Exynos 5422}, the CPU clusters and GPU on \textit{Kirin 970} (exclude NPU), and all the components on \textit{Kirin 970} (include NPU).
Overall, the co-execution energy efficiency is always better than the \textit{Big} CPU cluster. In \textit{Kirin 970} SoC, as GPU is much more energy-efficient than the CPU clusters, the co-execution provides better energy efficiency than the power-efficient \textit{Small} CPU cluster.

\section{Summary}
Mobile inferencing is now ubiquitous. In this work, we examine the power-performance characteristics of inferencing through several prominent neural networks on different components available within a mobile SoC. We also perform roofline analysis of networks on components to unveil the further optimization scope. We show that network throughput can increase by up to 2x using co-execution that engages all the components in inferencing simultaneously.

\vspace{-0.5in}
\begin{IEEEbiographynophoto}{Siqi Wang}
	is currently a research assistant and is working toward the Ph.D. degree at School of Computing, National University of Singapore. Her current research interests include performance optimization, task scheduling, general purpose GPUs and deep learning on heterogeneous multi-processor systems.
\end{IEEEbiographynophoto}
\vspace{-0.5in}
\begin{IEEEbiographynophoto}{Anuj Pathania} 
is currently working as a research fellow at School of Computing, National University of Singapore. He received his Ph.D. degree from Karlsruhe  Institute  of  Technology (KIT), Germany in 2018. His research focuses on resource management algorithms with emphasis on performance-, power- and thermal-efficiency  in  embedded  systems.
\end{IEEEbiographynophoto}
\vspace{-0.5in}
\begin{IEEEbiographynophoto}{Tulika Mitra}
	is a Professor of Computer Science at School of Computing, National University of Singapore. She received her PhD degrees in computer science from the State University of New York Stony Brook in 2000. Her research interests span various aspects of the design automation of embedded real-time systems, cyber-physical systems, and Internet of Things. 
\end{IEEEbiographynophoto}

\bibliographystyle{IEEEtran}
\bibliography{IEEEabrv,references}

\begin{thebibliography}{10}
\providecommand{\url}[1]{#1}
\csname url@samestyle\endcsname
\providecommand{\newblock}{\relax}
\providecommand{\bibinfo}[2]{#2}
\providecommand{\BIBentrySTDinterwordspacing}{\spaceskip=0pt\relax}
\providecommand{\BIBentryALTinterwordstretchfactor}{4}
\providecommand{\BIBentryALTinterwordspacing}{\spaceskip=\fontdimen2\font plus
\BIBentryALTinterwordstretchfactor\fontdimen3\font minus
  \fontdimen4\font\relax}
\providecommand{\BIBforeignlanguage}[2]{{%
\expandafter\ifx\csname l@#1\endcsname\relax
\typeout{** WARNING: IEEEtran.bst: No hyphenation pattern has been}%
\typeout{** loaded for the language `#1'. Using the pattern for}%
\typeout{** the default language instead.}%
\else
\language=\csname l@#1\endcsname
\fi
#2}}
\providecommand{\BIBdecl}{\relax}
\BIBdecl

\bibitem{wu2019machine}
C.-J. Wu, D.~Brooks, K.~Chen, D.~Chen, S.~Choudhury, M.~Dukhan, K.~Hazelwood,
  E.~Isaac, Y.~Jia, B.~Jia \emph{et~al.}, ``Machine learning at facebook:
  Understanding inference at the edge,'' in \emph{2019 IEEE International
  Symposium on High Performance Computer Architecture (HPCA)}.\hskip 1em plus
  0.5em minus 0.4em\relax IEEE, 2019, pp. 331--344.

\bibitem{wess2018weighted}
M.~Wess, S.~M.~P. Dinakarrao, and A.~Jantsch, ``Weighted
  quantization-regularization in dnns for weight memory minimization toward hw
  implementation,'' \emph{IEEE Transactions on Computer-Aided Design of
  Integrated Circuits and Systems}, vol.~37, no.~11, pp. 2929--2939, 2018.

\bibitem{pmu}
``{Keysight Technologies B2900 Series Precision Source/Measure Unit},''
  \url{https://goo.gl/U4HMbu}.

\bibitem{alexnet}
A.~Krizhevsky, I.~Sutskever, and G.~E. Hinton, ``Imagenet classification with
  deep convolutional neural networks,'' in \emph{Advances in neural information
  processing systems}, 2012, pp. 1097--1105.

\bibitem{googlenet}
C.~Szegedy, W.~Liu, Y.~Jia, P.~Sermanet, S.~Reed, D.~Anguelov, D.~Erhan,
  V.~Vanhoucke, and A.~Rabinovich, ``{Going deeper with convolutions},'' in
  \emph{Proceedings of the IEEE conference on computer vision and pattern
  recognition}, 2015, pp. 1--9.

\bibitem{mobilenet}
A.~G. Howard, M.~Zhu, B.~Chen, D.~Kalenichenko, W.~Wang, T.~Weyand,
  M.~Andreetto, and H.~Adam, ``{Mobilenets: Efficient convolutional neural
  networks for mobile vision applications},'' \emph{arXiv preprint:1704.04861},
  2017.

\bibitem{resnet}
K.~He, X.~Zhang, S.~Ren, and J.~Sun, ``{Deep residual learning for image
  recognition},'' in \emph{Proceedings of the IEEE conference on computer
  vision and pattern recognition}, 2016, pp. 770--778.

\bibitem{squeezenet}
F.~N. Iandola, S.~Han, M.~W. Moskewicz, K.~Ashraf, W.~J. Dally, and K.~Keutzer,
  ``{SqueezeNet: AlexNet-level accuracy with 50x fewer parameters and 0.5 MB
  model size},'' \emph{arXiv preprint :1602.07360}, 2016.

\bibitem{roofline}
S.~Williams, A.~Waterman, and D.~Patterson, ``{Roofline: An insightful visual
  performance model for floating-point programs and multicore architectures},''
  Lawrence Berkeley National Lab.(LBNL), Berkeley, CA (United States), Tech.
  Rep., 2009.

\bibitem{tinymembench}
S.~Siamashka, ``Tinymembench,'' \url{https://github.com/ssvb/tinymembench}.

\bibitem{pipe-it}
\BIBentryALTinterwordspacing
S.~Wang, G.~Ananthanarayanan, Y.~Zeng, N.~Goel, A.~Pathania, and T.~Mitra,
  ``High-throughput cnn inference on embedded arm big.little multi-core
  processors,'' \emph{IEEE Transactions on Computer-Aided Design of Integrated
  Circuits and Systems}, 2019. [Online]. Available:
  \url{http://dx.doi.org/10.1109/TCAD.2019.2944584}
\BIBentrySTDinterwordspacing

\end{thebibliography}
\end{document}